\documentclass[11pt,a4paper]{article}
\usepackage[hyperref]{emnlp2020}
\usepackage{times}
\usepackage{latexsym}
\usepackage{amsmath}
\usepackage{amssymb}
\usepackage{graphicx}
\usepackage{boldline}
\usepackage{multirow}
\usepackage{fixltx2e}
\usepackage{mathtools}
\usepackage{commath}
\usepackage{url}

\let\norm\undefined
\DeclarePairedDelimiter\norm{\lVert}{\rVert}

\aclfinalcopy 

\title{On Posterior Collapse and Encoder Feature Dispersion in Sequence VAEs}

\author{
  Teng Long$^1$, Yanshuai Cao$^1$, Jackie Chi Kit Cheung$^{1,2,3}$ \\
  $^1$Borealis AI \\
  $^2$McGill University, Mila \\
  $^3$Canada CIFAR AI Chair \\
  {\tt \{leo.long, yanshuai.cao\}@borealisai.com} \\
  {\tt jcheung@cs.mcgill.ca} \\
}
\date{}

\begin{document}
\maketitle

\begin{abstract}


Variational autoencoders (VAEs) hold great potential for modelling text, as they could
in theory separate high-level semantic and syntactic properties from local regularities
of natural language. Practically, however, VAEs with autoregressive decoders often suffer
from posterior collapse, a phenomenon where the model learns to ignore the latent variables,
causing the sequence VAE to degenerate into a language model. In this paper, we argue that
posterior collapse is in part caused by the lack of dispersion in encoder features. We provide
empirical evidence to verify this hypothesis, and propose a straightforward fix using pooling.
This simple technique effectively prevents posterior collapse, allowing model to achieve
significantly better data log-likelihood than standard sequence VAEs. Comparing to existing
work, our proposed method is able to achieve comparable or superior performances while being
more computationally efficient.

\end{abstract}

\section{Introduction}


Variational autoencoders (VAEs) \cite{kingma2013auto} are a class of latent-variable
models that allow tractable sampling through the decoder network and efficient
approximate inference via the encoder recognition network. \citet{bowman2016generating} proposed
an adaptation of VAEs for text in the hope that the latent variables could capture global
features while the decoder RNN can model the low-level local semantic and syntactic structures.
VAEs have been applied to many NLP-related tasks, such as language modeling, question answering
\cite{miao2016neural}, text compression \cite{miao2016language}, semi-supervised text
classification \cite{xu2017variational}, controllable language generation \cite{hu2017toward},
and dialogue response generation \cite{wen2017latent, zhao2017learning, park2018hierarchical}.
However, sequence VAE training can be brittle in practice; the latent variable is often
ignored while the model degenerates into a regular language model. This phenomenon occurs when the
inferred posterior distribution collapses onto the prior and is commonly referred to as
{\it posterior collapse} \cite{bowman2016generating}.

Previous work trying to address posterior collapse mostly falls into two categories. The first line
of work analyzes the problem from an optimization perspective \cite{alemi2018fixing} and proposes to
solve the issue with improved optimization schemes \cite{he2019lagging, liu2019cyclical}. The
other one focuses on architectural designs for different model components, for instance by intentionally
weakening the decoders \cite{semeniuta2017hybrid, yang2017improved}. However, these new optimization
methods usually come with hefty computation costs. And with weaker decoders, we are tackling
posterior collapse at the expense of losing on the expressive power of sequence VAE models.

In this paper, we analyze the issue from the perspective of the encoder. We argue that
posterior collapse is caused in part by the lack of dispersion in the deterministic features
produced by the
encoder. Input representations that are close to each other in feature space would lead to
approximate posteriors for each sequence concentrating in a small region, which makes latent
codes for different inputs somewhat indistinguishable. In the most extreme case, all encoder
features would collapse onto a single point, thus there would be no mutual information between
the input sequences and their latent variables. During training, since latent variables carry
no information to help the decoder better reconstruct the input, optimization would push
approximate posteriors to prior in
order to avoid paying the cost of the KL term in the ELBO objective, leading to posterior collapse.

We provide empirical evidence that posterior collapse in text VAEs is causally related to the lack
of dispersion in the encoder features. Furthermore, we propose to use simple pooling operations
instead of the last hidden state of the encoder to alleviate posterior collapse without major
modifications to the standard VAE formulation and optimization. Our contributions are three-fold.
\textit{1)} We analyze posterior collapse from a different angle than existing work, namely the
lack of feature dispersion. \textit{2)} We present empirical evidence to support our hypothesis.
\textit{3)} We take it as motivation to propose a simple method leveraging pooling operations
to address posterior collapse. This simple technique can effectively prevent posterior collapse
and achieve significantly better log-likelihood than standard sequence VAE models. When comparing
to existing methods, our proposed method is able to achieve either comparable or superior results
while being more compurationally efficient.


\section{Related Work}
\label{related_work}

Prior work that aim to address posterior collapse roughly fall into the following
two categories. The first line of work tries to analyze this issue from the optimization
perspective. The other one focuses on the architectural design of the model.

\citet{bowman2016generating} initially proposed to use a simple annealing
schedule that starts with a small value and gradually increases to 1 for the KL term
in the ELBO objective at the beginning of training. However in practice, this trick along
is not sufficent to prevent posterior collapse. Later, \citet{higgins2017beta} proposed
$\beta$-VAEs, for which the weight for KL term is considered as a hyperparameter and is
usually set to be smaller than 1. Doing so could generally avoid posterior collapse,
but at the cost of worse NLLs. More recently, \citet{liu2019cyclical}
proposed cyclical annealing schedule, which repeats the annealing process
multiple times in order to help optimization to escape bad local minima.
\citet{he2019lagging} argued that posterior collapse is caused by the approximate
posterior $q_{\phi}(z|x)$ lagging behind the intractable true posterior
$p_{\theta}(z|x)$ during training, and thus proposed to always train the encoder
till convergence before updating to the decoder. In this work, we add another
perspective for analyzing the issue.

Previously proposed architectural changes mainly focus on the decoder network
and the choice of the approximate posteriors. \citet{semeniuta2017hybrid}
and \citet{yang2017improved} argued that posterior collapse was caused by powerful
autoregressive decoders and proposed to intentionally weaken the decoder, forcing it
to rely more on the latent variables to reconstruct the input,
which also leads to worse estimated data likelihood.
\citet{dieng2018avoiding} proposed to add skip connections from the latent variables
to lower layers of the decoder and proved that doing so increases the mutual information
between data and latent codes.
On the other hand, \citet{kim2018semi},
\citet{xu2018spherical} and \citet{razavi2019preventing} argued that using multivariate Gaussian
is inherently flawed and advocated for augmenting the amortized approximate posteriors
with instance-based inference, or using completely different probability distributions
for both the prior and the approximate posterior. Additionally, \citet{wang2019riemannian}
tried to address the limitation of the Gaussian assumption by transforming the latent
variables with flow-based models and minimizing the Wasserstein distance between the
marginal distribution and the prior directly.

\section{Preliminaries}

\subsection{VAE Formulation}
\label{vae_formulation}

Variational autoencoders were initially proposed by \citet{kingma2013auto}. Compared
to the standard autoencoders, VAEs introduce an explicitly parameterized latent variable
$z$ over data $x$. Instead of directly maximizing the log likelihood of data, VAEs are
trained to maximize to the Evidence Lower Bound (ELBO) on log likelihood:
\begin{align*}
    \log p_{\theta}(x)
    &\geq \mathbb{E}_{q_{\phi}(z|x)}[\log p_{\theta}(x|z)] \\
    &\phantom{\geq_{\theta}} - D_{KL}(q_{\phi}(z|x)|p_{\theta}(z)) \\
    &= \mathcal{L}(\theta, \phi; x)
\end{align*}
where $p_{\theta}(z)$ is the prior distribution, $q_{\phi}(z|x)$ is typically
referred to as the recognition model (also known as the encoder), and
$p_{\theta}(x|z)$ is the generative model (also known as the decoder).

The ELBO objective $\mathcal{L}(\theta, \phi; x)$ consists of two terms. The first
one is the reconstruction term $\mathbb{E}_{q_{\phi}(z|x)}[\log p_{\theta}(x|z)]$, which trains
the generative model to reconstruct input data $x$ given its latent variable $z$. The second
term is the KL divergence to $q_{\phi}(z|x)$ from $p_{\theta}(z)$, which
penalizes the approximate posteriors produced by recognition model for deviating from the prior
too much.

In standard VAEs, the prior is typically assumed to be the isotropic Gaussian; i.e.,
$p_{\theta}(z)=\mathcal{N}(\mathbf{0}, \mathbf{I})$. The approximate posterior for
$x$ is defined as a multivariate Gaussian with diagonal covariance
matrix whose parameters are functions of $x$, thus
$q_{\phi}=\mathcal{N}(\mu_{\phi}(x), \sigma_{\phi}^{2}(x))$ with $\phi$ being the
parameters of recognition model. Such assumptions ensure that the forward and
backward passes can be performed efficiently during training, and the KL term
can be computed analytically.

\subsection{Sequence VAEs}

Inspired by \citet{kingma2013auto}, \citet{bowman2016generating} proposed an
adaptation of variational autoencoders for generative text modeling, dubbed the
Sequence VAEs (SeqVAEs). Neural language models
typically predict each token $x_{t}$ conditioned on the history of previously generated
tokens:
\begin{equation*}
    p(x) = \prod_{t=1}^{T} p(x_{t}|x_{1}, x_{2}, ..., x_{t - 1})
\end{equation*}

Rather than directly modeling the above factorization of sequence $x$,
\citet{bowman2016generating} specified a generative process for input sequence
$x$ that is conditioned on some latent variable $z$:
\begin{equation*}
    p(x|z) = \prod_{t=1}^{T} p(x_{t}|x_{1}, x_{2}, ..., x_{t- 1}, z)
\end{equation*}
where the marginal distribution $p(x)$ could in theory be recovered by integrating
out the latent variable. The hope is that latent variable $z$ would be able to
capture certain holistic properties of the input sentences, such as their topics
and styles.

Autoregressive architectures such
as RNNs are the ideal choice for parameterizing the encoder and the decoder
in SeqVAEs. Specifically, the encoder first reads the entire sentence $x$ in order
to produce feature vector $h$ for the sequence. The feature vector is then fed to
some linear transformation to produce the mean and covariance of approximate
posterior. A latent code $z$ is sampled from the approximate posterior and then
passed to the decoder network to reconstruct input $x$.

\subsection{Posterior Collapse}

An alternative interpretation for VAEs is to view them as a regularized version of
the standard autoencoders. The reconstruction term in the ELBO
objective encourages the latent code $z$ to convey meaningful information in order
to reconstruct $x$. On the other hand, the KL divergence term
penalizes $q_{\phi}(z|x)$ for deviating from $p_{\theta}(z)$ too much,
preventing the model from simply memorizing each data point.
This creates the possibility of an undesirable local optimum in which the
approximate posterior becomes nearly identical to the prior distribution,
i.e. $q_{\phi}(z|x) \approx p_{\theta}(z)$ for all $x$.

Such a degenerate solution is commonly known as \textit{posterior collapse} and is
often signalled by the close-to-zero KL term in the ELBO objective during training. When
optimization reaches the collapsed solutions, the approximate posterior resembles
the prior distribution and conveys no useful information about the corresponding
data $x$, which defeats the purpose of having a recognition model. In this case,
the decoder would have no other choice but to ignore the latent codes.

Posterior collapse is particularly prevalent when applying VAEs to text.
To address the issue,
\citet{bowman2016generating} proposed to gradually increase weight of the
KL regularizer from a small value to $1$ following a simple annealing schedule.
However, in practice, this method alone is not sufficient to prevent posterior
collapse \cite{xu2017variational}.

\section{The Importance of Feature Dispersion}

\begin{figure*}
    \centering
    \includegraphics[scale=0.38]{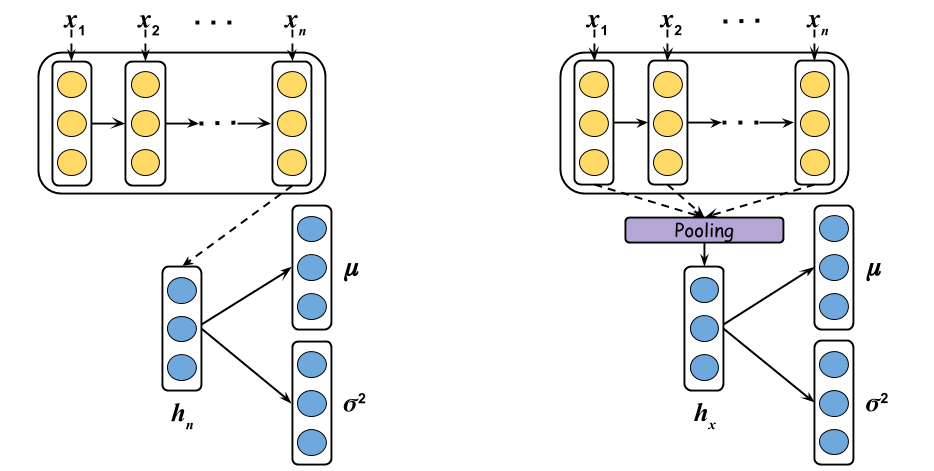}
    \caption{\textbf{Left:} The typical architecture of recognition model for a
    sequence VAE in which only the last hidden state $h_n$ from encoder RNN is used
    as feature representation to compute the mean $\mu$ and variance $\sigma^{2}$
    parameters of approximate posteriors $q_{\phi}(z|x)$. \textbf{Right:} Our proposed
    modification of how the feature vector $h_x$ for sequence $x$ is computed.
    Specifically, $h_x$ is now computed by performing pooling over the temporal
    dimension of all hidden states $h=[h_1,h_2,...,h_n]$ output by RNN encoder, which
    is then used to compute parameters $\mu$ and $\sigma^{2}$ as usual.}
    \label{fig:comparison}
\end{figure*}

\subsection{Issues with Last Hidden States}
\label{last_hid}

In sequence VAEs, the encoder RNN processes the input sentence $x=[x_1,x_2,...,x_n]$
one word at a time to produce a series of hidden states $h=[h_1,h_2,...,h_n]$. In
the typical architecture, the last hidden state $h_n$ is taken as the feature
to compute the mean and variance for the approximate posterior,
as shown on the left side of Figure \ref{fig:comparison}, thus:
\begin{align*}
    q_{\phi}(z|x)=&\mathcal{N}(\mu_{\phi}(x), \sigma_{\phi}^{2}(x)) \\
    \textbf{s.t.} \phantom{\theta} \mu_{\phi}(x)&=W_{1} * h_n + b_{1} \\
                  \phantom{\theta} \sigma_{\phi}^{2}(x)&=\textit{exp}(W_{2} * h_n + b_{2})
\end{align*}
where $W_{1}$, $b_{1}$ and $W_{2}$, $b_{2}$ are the linear layer parameters for mean and
log-variance respectively.

However, using the last hidden states as feature representations could be problematic, as
RNNs are known to have issues retaining information further back in history. A a result,
$h_n$ tends to be dominated by later words in the input. We hypothesize that such tendencies
of RNNs would result in a feature space with insufficient dispersion.

In the most extreme case, all encoder features would collapse onto a single point regardless
of its input, thus there would be no mutual information between the input sequences and their
latent variables, which implies $q_{\phi}(z|x)=q_{\phi}(z)$. In practice, features
from a space with insufficient dispersion would result in approximate posteriors concentrating
in a small region of posterior space, with high chances of overlap for different input data.

As a result, latent codes sampled from different approximate posteriors would look somewhat
similar, thus provide little useful information to the decoder. Since no useful information
could be conveyed by the latent variables, optimization would push approximate posteriors
towards the prior to avoid paying the KL penalty and maximize the overall ELBO objective,
thus causing training to reach the undesirable local optimum that is posterior collapse.
Therefore we argue that maintaining dispersion of the feature space is important to prevent
posterior collapse.




\subsection{Increasing Dispersion Reduces Collapse}
\label{visualize}

In order to verify the aforementioned intuition, we train two different sequence VAE models whose
encoders are parameterized by LSTMs with only three hidden units and with three-dimensional latent
variables on Yahoo dataset \cite{yang2017improved}. Although it is clearly not the optimal
configurations for the encoder and latent variables, doing so would enable us to visualize
the feature space produced by the encoders explicitly.

The first model follows standard sequence
VAE settings and is trained to optimize the ELBO objective. The other one is trained with an additional
Cosine Regularizer that minimizes the pairwise cosine similarities of encoder features within each batch;
i.e., $cos(h_{n}^{i}, h_{n}^{j}) = \frac{h_{n}^{i} \cdot h_{n}^{j}}{\norm{h_{n}^{i}} \norm{h_{n}^{j}}}$
where $h_{n}^{i}$ and $h_{n}^{j}$ are feature vectors for the $i$-th and $j$-th sequences in the batch,
in order to examine whether explicitly encouraging feature dispersion would reduce the chance of posterior
collapse. Other decorrelation or pairwise repulsion regularizations were explored under different contexts
\cite{cogswell2015reducing, cao2018improving}


Figure \ref{fig:yahoo_3d} a) and b) visualize each feature space on the validation set. Notice that the standard
sequence VAE maps all sentences to a concentrated region in feature space. On the other hand, the model
trained with the feature cosine regularizer generates more dispersed features. As a result, we can
see from Figure \ref{fig:toy_kld} that the KL term on validation set quickly converges to close to
zero for the vanilla sequence VAE, while the KL term plateaus at non-zero values for model trained
with the cosine regularizer. This confirms our intuition that posterior collapse is caused in part
by the lack of dispersion in features from the encoder, which in turn result in nearly indistinguishable
latent codes for different input sequences.


\subsection{Achieving Better Dispersion via Pooling}
\label{pooling}


\begin{figure*}
    \centering
    \includegraphics[scale=0.375]{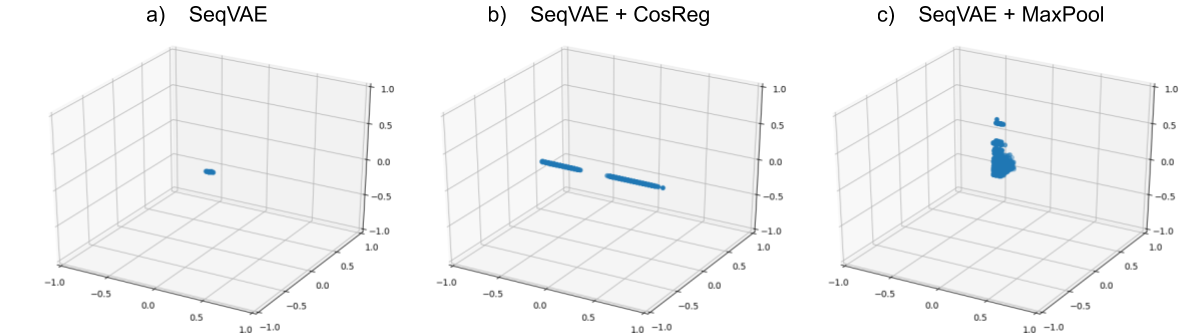}
    \caption{Feature space visualizations on Yahoo validation set for vinilla sequence VAE,
    sequence VAE with feature cosine regularizer, and sequence VAE with max pooling.}
    \label{fig:yahoo_3d}
\end{figure*}

\begin{figure}
    \centering
    \includegraphics[scale=0.45]{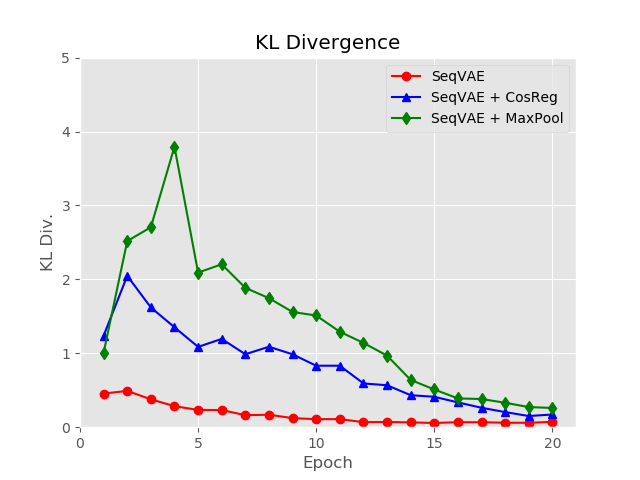}
    \caption{KL diverence on Yahoo validation set (three-dimensional latent variables).}
    \label{fig:toy_kld}
\end{figure}


In practice, however, the feature cosine regularizer is not ideal as it can be difficult to optimize
for pairwise repulsion objectives in high dimensional space, resuling in inferior model performance.
Following our intuition, we want to find better alternatives to generate dispersed features for
$x=[x_1,x_2,...,x_n]$. Ideally, we would want to make use of information across all hidden states
instead of just the last step. Thus, we would like the feature vector $h_{x}$ for sequence $x$ to be:
\begin{equation*}
    h_x=\textit{aggregate}([h_1,h_2,...,h_n])
\end{equation*}
where $\textit{aggregate}$ is some function that takes a list of vectors and produces a single feature
vector.


To avoid adding more parameters to the model, we choose to experiment with different \textit{pooling}
functions. Since attention mechanism is the prevalent method for feature aggregation in many NLP-related
tasks, pooling is not as widely used in NLP as in computer vision. However, there have also been
successful applications of pooling in NLP, such as multi-task learning \cite{collobert2008unified}
and learning pretrained universal sentence encoders \cite{conneau2017supervised}.

In the context of sequence VAE models, we perform pooling over the temporal dimension of
hidden states $h=[h_1,h_2,...,h_n]$ produced by the encoder RNN, as illustrated on the
right side of Figure \ref{fig:comparison}. We experiment with \textit{three} options,
the first two are the commonly used \textit{average pooling (AvgPool)} and
\textit{max pooling (MaxPool)}. The last one performs max pooling based on the absolute
values of each element while preserving the signs of the pooled elements, which we refer
to as \textit{sign-preserved absolute pooling (AbsPool)}.


There are also other alternatives for the aggregate function. One option is to jointly
learn a self-attention module to perform the aggregation \cite{yang2016hierarchical}.
We experimented with this approach and found it to be outperformed by pooling-based
methods. We suspect that it could be due to the fact that the attention module adds
additional parameters to the model and causes it to overfit more easily, thus creating
more complications to the already challenging optimization problem.

To examine the effect of pooling on the encoder features, we follow the setup in Section
\ref{visualize} and train another model of the same size equipped with max pooling. As
shown in Figure \ref{fig:yahoo_3d} c), pooling is able to increase the dispersion
in the feature space even more compared to the cosine regularizer. As a result, we can see
that the KL term plateaus at higher values in Figure \ref{fig:toy_kld}, which again aligns
with our claim that having dispersed features is causually related to avoiding posterior collapse.



\section{Experiments}



In this section, we present the main experimental results on benchmark datasets. We
also run additional experiments in order to gain more insights into different methods.

\subsection{Settings}

\begin{figure*}[!ht]
    \centering
    \includegraphics[scale=0.30]{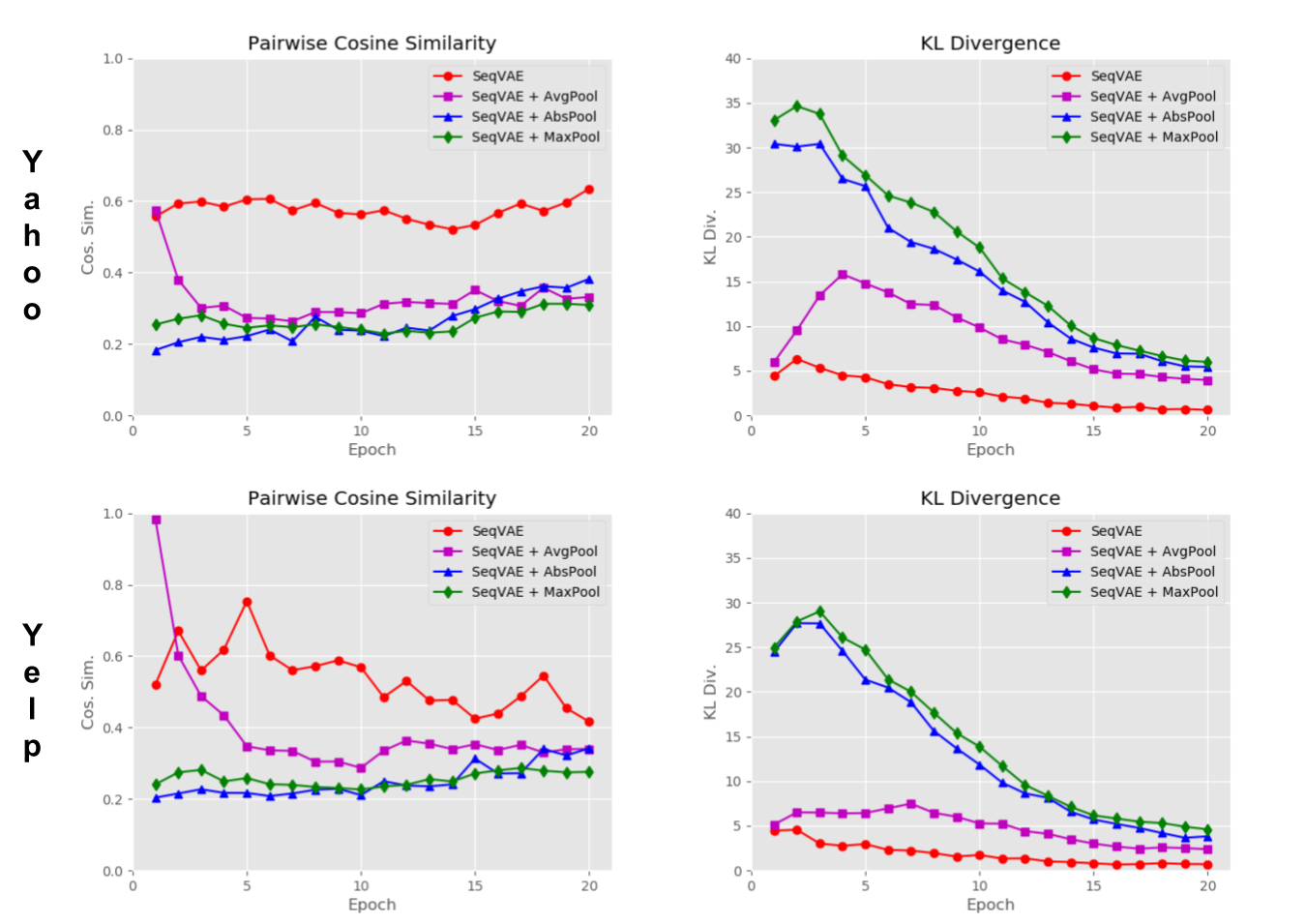}
    \caption{Pairwise cosine similarities between feature vectors and KL divergences
    for the validation sets. Notice that for regular SeqVAE models, the cosine
    similarities among different sequences remain at a higher level as the training
    progresses. At a result, the KL term quickly collapses to close to zero. On the
    other hand, pooling is able to maintain the dispersion in the feature space,
    thus helping to avoid posterior collapse.}
    \label{fig:cos_kld}
\end{figure*}

We evaluate all models on two benchmark datasets for text VAEs: \textit{Yahoo} and \textit{Yelp}
\cite{yang2017improved}. Both datasets consist of train, valid, and test splits of \textit{100k},
\textit{10k}, and \textit{10k} sentences, with the average lengths of \textit{78.76} for Yahoo and
\textit{96.01} for Yelp.

Following the experiment settings from previous work \cite{kim2018semi,he2019lagging},
we employ single-layer LSTMs with 1024 hidden units for both the encoder and decoder
with a latent space of 32 dimensions. For all models, we use the isotropic Gaussian
$\mathcal{N}(\textbf{0}, \textbf{I})$ as prior and the recognition model parameterizes
a multivariate Gaussian with diagonal covariance matrix. We train the model standard SGD
with early stopping.


\subsection{Analysis of Feature Dispersion}

For larger models, since we are unable to visualize high dimensional space without losing
information, we monitor the pairwise cosine similarities of feature vectors for sequences
from the validation set. Figure \ref{fig:cos_kld} shows the average pairwise cosine similarities
and average KL divergence for both benchmark datasets during training.


Observe that for the regular sequence VAE, the average pairwise cosine similarities on the
validation set remain at a higher level compared to the pooling-based models. As the training
progresses, KL divergence is quickly pushed to take on small values by the optimization and
gradually approach zero, signalling the occurrence of posterior collapse. Whereas for models
equipped with pooling, the cosine similarities are kept at a lower level, indicating more
dispersed and diverse feature space. As a result, KL terms for pooling-based models converge
to non-zero values.


\subsection{Quantitative Results}
\label{sec:results}

\begin{table*}
    \centering
    \begin{tabular}{ccccccccccc}
    \hlineB{3}
    \\ [-2.25ex]
    \multicolumn{1}{c|}{}                          & \multicolumn{5}{c}{\textbf{Yahoo}}                                                         & \multicolumn{5}{|c}{\textbf{Yelp}}                                                                              \\
    \multicolumn{1}{c|}{\textbf{Model}}            & \textbf{NLL}   & $\overline{\textbf{NLL}}$   & \textbf{KL}   & \textbf{MI}   & \textbf{AU} & \multicolumn{1}{|c}{\textbf{NLL}}   & $\overline{\textbf{NLL}}$   & \textbf{KL}   & \textbf{MI}   & \textbf{AU} \\
    \\ [-2.25ex]
    \hlineB{3}
    \\ [-2.25ex]
    \multicolumn{1}{l|}{LSTM-LM*}             & 328.0           & --  & --   & --  & -- & \multicolumn{1}{|c}{358.1}           & --  & --   & --   & -- \\
    \multicolumn{1}{l|}{SeqVAE}               & 328.6           & --  & 0.0  & 0.0 & 0  & \multicolumn{1}{|c}{358.1}           & --  & 0.3  & 0.3  & 1  \\
    \multicolumn{1}{l|}{SeqVAE + WordDrop}    & 330.7           & --  & 5.4  & 3.0 & 6  & \multicolumn{1}{|c}{362.2}           & --  & 1.0  & 0.8  & 1  \\
    \multicolumn{1}{l|}{SkipVAE}              & 328.1           & --  & 4.5  & 2.4 & 11 & \multicolumn{1}{|c}{357.4}           & --  & 2.5  & 1.5  & 4  \\
    \multicolumn{1}{l|}{WAE-RNF**}            & 339.0           & --  & 3.0  & --  & -- & \multicolumn{1}{|c}{--}              & --  & --   & --   & -- \\
    \multicolumn{1}{l|}{SeqVAE + Cyclical}    & 328.6           & --  & 0.0  & 0.0 & 0  & \multicolumn{1}{|c}{358.4}           & --  & 0.4  & 0.3  & 1  \\
    \multicolumn{1}{l|}{SeqVAE + Aggressive}  & \textbf{326.7}  & --  & 5.7  & 2.9 & 15 & \multicolumn{1}{|c}{\textbf{355.9}}  & --  & 3.8  & 2.4  & 11 \\
    \\ [-2.25ex]
    \hlineB{2}
    \\ [-2.25ex]
    \multicolumn{1}{l|}{SeqVAE + AvgPool}      & 327.8          & --  & 2.4 & 1.6 & 5  & \multicolumn{1}{|c}{357.5}          & --  & 1.6 & 1.2 & 5  \\
    \multicolumn{1}{l|}{SeqVAE + AbsPool}      & 327.4          & --  & 3.6 & 2.4 & 8  & \multicolumn{1}{|c}{356.6}          & --  & 2.0 & 1.7 & 7  \\
    \multicolumn{1}{l|}{SeqVAE + MaxPool}      & 327.2          & --  & 3.7 & 2.5 & 9  & \multicolumn{1}{|c}{356.0}          & --  & 3.1 & 2.2 & 8  \\
    \\ [-2.25ex]
    \hlineB{2}
    \\ [-2.25ex]
    \multicolumn{1}{l|}{iVAE}                   & --  & 309.5 & 8.0  & 4.4  & 32 & \multicolumn{1}{|c}{--} & 348.2  & 7.6  & 4.6  & 32 \\
    \\ [-2.25ex]
    \hlineB{3}
    \end{tabular}
    \caption{Experiment results on the \textit{Yahoo} and \textit{Yelp} datasets. For
    the LSTM-LM*, we report the exact negative log likelihood. For the WAE-RNF**, we
    only show the results on \textit{Yahoo} reported by \citet{wang2019riemannian} as
    their experiments on \textit{Yelp} were conducted on a different version of the
    dataset. Note that the estimated negative log likelihood from iVAE \citep{fang2019implicit}
    cannot be directly compared with other methods.}
    \label{tab:results}
\end{table*}


\begin{table}[!ht]
    \centering
    \begin{tabular}{ccc}
    \hlineB{3}
    \\ [-2.25ex]
    \multirow{2}{*}{} & \multicolumn{1}{c}{\textbf{Yahoo}} & \multicolumn{1}{c}{\textbf{Yelp}} \\
                      & \textbf{Updates}                   & \textbf{Updates} \\
    \\ [-2.25ex]
    \hlineB{3}
    \\ [-2.25ex]
    \multicolumn{1}{l|}{SeqVAE + Aggressive} & 608k & 625k \\
    \multicolumn{1}{l|}{SeqVAE + MaxPool}    & 199k & 196k \\
    \\ [-2.25ex]
    \hlineB{3}
    \end{tabular}
    \caption{Computation costs of aggressive training \textit{vs} max pooling,
    measured in terms of parameter updates.}
    \label{tab:compute_cost}
\end{table}

We also compare quantitatively with other existing methods on the benchmark datasets. We report
approximate negative log likelihood (\textbf{NLL}) estimated by 500 importance weighted samples
\cite{burda2015importance}. We also report KL divergence $D_{KL}(q_{\phi}(z|x)|p(z))$ (\textbf{KL}),
estimated mutual information (\textbf{MI}) between $x$ and $z$ \cite{dieng2018avoiding}, and number
of active units (\textbf{AU}) in the latent codes \cite{burda2015importance}.


Note that although metrics such as KL, MI, and AU provide certain insights for the
models; i.e., whether posterior collapse has occurred for a particular model, they do
not directly correlate with the overall qualitiy of a model. Thus higher KL divergence
or numbers of active units are not necessarily better, as illustrated by our results.
Ultimately, what we want from a model is lower negative log likelihood (with non-zero
KL divergence) since it is a direct indicator of how well our models capture the data
distribution.

We compare our models with the following methods from the literature: SkipVAE
\cite{dieng2018avoiding}, WAE-RNF \cite{wang2019riemannian} which make modifications to
decoders or variational distributions; and Cyclical Annealing \cite{liu2019cyclical},
Aggressive Training \cite{he2019lagging} which aim to prevent posterior collapse with
new optimization schemes. Aggressive training in particular comes with very high computation
costs since it requires to train the encoder to near convergence before each decoder update.
Additionally, we train two baseline sequence VAE models: one only with KL annealing; and the
other with both KL annealing and Word Dropout as in \citet{bowman2016generating}. All models,
including different models from the literature, are trained following a simple linear KL
annealing schedule at early stage of training except for ones trained with cyclical
annealing schedule.


\citet{fang2019implicit} proposed to use implicit distributions as the approximate posteriors,
for which they named their model implicit VAE (iVAE). At the first glance, their model improved
upon the previous methods by a large margin. Howerver, their claimed results $\overline{\textbf{NLL}}$
are in fact a lower bound on the true NLL of the data and thus cannot be directly compared to the
results of other models (see Section A in the appendix for more details).



Table \ref{tab:results} shows the quantitative results. We observe that
pooling can effectively prevent posterior collapse while achieving significantly lower
estimated NLLs compared to standard sequence VAEs, with max pooling offering the best
performances on both datasets. This suggests that increasing feature dispersion can
effectively prevent posterior collapse, which leads to better overall modelling qualitiy.
Applying heavy word dropout leads to non-zero KL term but also worse log likelihood. This
indicates that solving posterior collapse is necessary but not sufficient to improve NLLs
in sequence VAE models.

Notice that although average pooling also improves upon the baseline model, it provide the
least amount of improvements compared to the other two pooling methods. The performance
gap is noticeably more significant on Yelp dataset where the average sequence length is
longer, which aligns with our intuition that average pooling is likely to produce less
dispersion over longer input sequences due to the central limit theorem.


We also see that our methods outperform both SkipVAE and WAE-RNF, which suggests
that certain proposed architectural changes might not be necessary to improve upon
the original sequence VAE models with Gaussian distributions. \citet{liu2019cyclical}
reported promising results for text modelling on the relatively simple Penn Tree
Bank (PTB) dataset with their proposed cyclical annealing schedule. However the
same success is not carried over to more complex data. Aggressive training gives
the best estimated NLLs on both datasets; our methods are able to achieve comparable
performances, particularly on the more challenging \textit{Yelp} dataset where the
average length is longer, while being significantly more computationally efficient,
as shown in Table \ref{tab:compute_cost}.

\subsection{Comparison with Aggressive Training}

\begin{figure}
    \centering
    \includegraphics[scale=0.45]{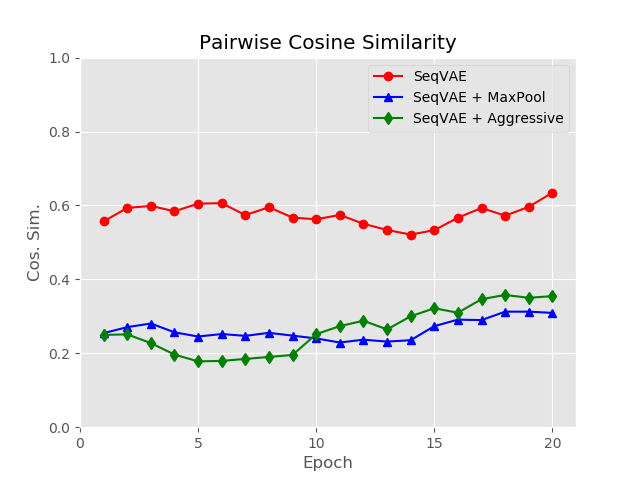}
    \caption{Pairwise cosine similarities between feature vectors on the validation set
    on Yahoo.}
    \label{fig:yahoo_cos}
\end{figure}

From Table \ref{tab:results}, we see that pooling and aggressive training are able to
offer much bigger improvements to the standard sequence VAE models compared to other
baseline models. To better understand the connections between these two methods, we
again monitor the pairwise cosine similarities averaged over the validation set as the
training progresses, which is illustrated in Figure \ref{fig:yahoo_cos}.

We observe that for both aggressive training and max pooling, the average pairwise
cosine similarities among feature representations produced by the encoder are kept
at a lower level as opposed to the baseline model, which suggests that the success
of aggressive training could also be attributed to the increase of dispersion in
feature space. The difference is that aggressive training achieves this with a new
optimization scheme whose computation cost is three times higher than our methods.
This further support our claim that posterior collapse in sequence VAEs is caused
in part by the lack of dispersion in feature space, and increasing dispersion could
prevent it from happening.


\subsection{How Important is KL Annealing?}

\begin{table}
    \centering
    \begin{tabular}{ccccc}
    \hlineB{3}
    \\ [-2.25ex]
    \multirow{2}{*}{} & \multicolumn{2}{c}{\textbf{Yahoo}} & \multicolumn{2}{c}{\textbf{Yelp}} \\
                      & \textbf{W.O.} & \textbf{With}      & \textbf{W.O.} & \textbf{With}     \\
    \\ [-2.25ex]
    \hlineB{3}
    \\ [-2.25ex]
    SkipVAE    & \multicolumn{1}{|c}{329.1} & 328.1 & \multicolumn{1}{|c}{358.2} & 357.4 \\
    Aggressive & \multicolumn{1}{|c}{328.2} & 326.7 & \multicolumn{1}{|c}{356.9} & 355.9 \\
    MaxPool    & \multicolumn{1}{|c}{328.6} & 327.2 & \multicolumn{1}{|c}{357.6} & 356.0 \\
    \\ [-2.25ex]
    \hlineB{3}
    \end{tabular}
    \caption{Estimated NLLs of various models when trained without \textit{vs} with KL
    annealing.}
    \label{tab:kl_anneal}
\end{table}

As mentioned previously, the experimental results of various models presented in Section
\ref{sec:results} were achieved with KL annealing, which is necessary to achieve the
best possible data log likelihood. As a matter of fact, it is often used together with
the proposed algorithms in order to achieve the best possible results. To illustrate
the importance of KL annealing, we compare the estimated NLLs of SkipVAE, Aggressive
Training, and MaxPool when trained without and with KL annealing.

As shown in Table \ref{tab:kl_anneal}, KL annealing is indeed rather important and
necessary if we want a model that better captures the data distribution. Note that
in most cases, the gap for estimated NLLs between whether using it or not is rather
significant, suggesting that KL annealing might be able to help the model to better
explore during early stage of learning and eventually reach better local optimum.
Additional research is needed to better understand the effects of KL annealing in
optimizing variational models and why it is so crucial for reaching a better local
optimum of ELBO.

\section{Conclusion}

In this paper, we analyze posterior collapse in sequence VAEs from the perspective of
the encoder network. We argue that the issue is caused in part by the lack of dispersion
in features from the encoder. We provide empirical evidence to verify this hypothesis and
propose a simple architectural change that utilizes pooling operations. Our proposed methods
can effectively prevent posterior collapse while achieving comparable or better NLLs compared
to existing methods without any additional computation costs.

\bibliography{references}

\begin{thebibliography}{27}
\expandafter\ifx\csname natexlab\endcsname\relax\def\natexlab#1{#1}\fi

\bibitem[{Alemi et~al.(2018)Alemi, Poole, Fischer, Dillon, Saurous, and
  Murphy}]{alemi2018fixing}
Alexander Alemi, Ben Poole, Ian Fischer, Joshua Dillon, Rif~A Saurous, and
  Kevin Murphy. 2018.
\newblock Fixing a broken elbo.
\newblock In \emph{International Conference on Machine Learning}, pages
  159--168.

\bibitem[{Bowman et~al.(2016)Bowman, Vilnis, Vinyals, Dai, Jozefowicz, and
  Bengio}]{bowman2016generating}
Samuel~R Bowman, Luke Vilnis, Oriol Vinyals, Andrew Dai, Rafal Jozefowicz, and
  Samy Bengio. 2016.
\newblock Generating sentences from a continuous space.
\newblock In \emph{Proceedings of The 20th SIGNLL Conference on Computational
  Natural Language Learning}, pages 10--21.

\bibitem[{Burda et~al.(2016)Burda, Grosse, and
  Salakhutdinov}]{burda2015importance}
Yuri Burda, Roger Grosse, and Ruslan Salakhutdinov. 2016.
\newblock Importance weighted autoencoders.
\newblock In \emph{International Conference on Learning Representations}.

\bibitem[{Cao et~al.(2018)Cao, Ding, Lui, and Huang}]{cao2018improving}
Yanshuai Cao, Gavin~Weiguang Ding, Kry Yik-Chau Lui, and Ruitong Huang. 2018.
\newblock Improving gan training via binarized representation entropy (bre)
  regularization.
\newblock In \emph{International Conference on Learning Representations}.

\bibitem[{Cogswell et~al.(2016)Cogswell, Ahmed, Girshick, Zitnick, and
  Batra}]{cogswell2015reducing}
Michael Cogswell, Faruk Ahmed, Ross Girshick, Larry Zitnick, and Dhruv Batra.
  2016.
\newblock Reducing overfitting in deep networks by decorrelating
  representations.
\newblock In \emph{International Conference on Learning Representations}.

\bibitem[{Collobert and Weston(2008)}]{collobert2008unified}
Ronan Collobert and Jason Weston. 2008.
\newblock A unified architecture for natural language processing: Deep neural
  networks with multitask learning.
\newblock In \emph{Proceedings of the 25th international conference on Machine
  learning}, pages 160--167. ACM.

\bibitem[{Conneau et~al.(2017)Conneau, Kiela, Schwenk, Barrault, and
  Bordes}]{conneau2017supervised}
Alexis Conneau, Douwe Kiela, Holger Schwenk, Lo{\"\i}c Barrault, and Antoine
  Bordes. 2017.
\newblock Supervised learning of universal sentence representations from
  natural language inference data.
\newblock In \emph{Proceedings of the 2017 Conference on Empirical Methods in
  Natural Language Processing}, pages 670--680.

\bibitem[{Dieng et~al.(2019)Dieng, Kim, Rush, and Blei}]{dieng2018avoiding}
Adji~B Dieng, Yoon Kim, Alexander~M Rush, and David~M Blei. 2019.
\newblock Avoiding latent variable collapse with generative skip models.
\newblock In \emph{Proceedings of the 22nd International Conference on
  Artificial Intelligence and Statistics (AISTATS 2019)}.

\bibitem[{Fang et~al.(2019)Fang, Li, Gao, Dong, and Chen}]{fang2019implicit}
Le~Fang, Chunyuan Li, Jianfeng Gao, Wen Dong, and Changyou Chen. 2019.
\newblock Implicit deep latent variable models for text generation.
\newblock In \emph{EMNLP}.

\bibitem[{He et~al.(2019)He, Spokoyny, Neubig, and
  Berg-Kirkpatrick}]{he2019lagging}
Junxian He, Daniel Spokoyny, Graham Neubig, and Taylor Berg-Kirkpatrick. 2019.
\newblock Lagging inference networks and posterior collapse in variational
  autoencoders.
\newblock In \emph{International Conference on Learning Representations}.

\bibitem[{Higgins et~al.(2017)Higgins, Matthey, Pal, Burgess, Glorot,
  Botvinick, Mohamed, and Lerchner}]{higgins2017beta}
Irina Higgins, Loic Matthey, Arka Pal, Christopher Burgess, Xavier Glorot,
  Matthew Botvinick, Shakir Mohamed, and Alexander Lerchner. 2017.
\newblock beta-vae: Learning basic visual concepts with a constrained
  variational framework.
\newblock In \emph{International Conference on Learning Representations},
  volume~3.

\bibitem[{Hu et~al.(2017)Hu, Yang, Liang, Salakhutdinov, and
  Xing}]{hu2017toward}
Zhiting Hu, Zichao Yang, Xiaodan Liang, Ruslan Salakhutdinov, and Eric~P Xing.
  2017.
\newblock Toward controlled generation of text.
\newblock In \emph{Proceedings of the 34th International Conference on Machine
  Learning-Volume 70}, pages 1587--1596. JMLR. org.

\bibitem[{Kim et~al.(2018)Kim, Wiseman, Miller, Sontag, and Rush}]{kim2018semi}
Yoon Kim, Sam Wiseman, Andrew Miller, David Sontag, and Alexander Rush. 2018.
\newblock Semi-amortized variational autoencoders.
\newblock In \emph{International Conference on Machine Learning}, pages
  2683--2692.

\bibitem[{Kingma and Welling(2014)}]{kingma2013auto}
Diederik~P Kingma and Max Welling. 2014.
\newblock Auto-encoding variational bayes.
\newblock In \emph{Proceedings of the 2nd International Conference on Learning
  Representations (ICLR)}.

\bibitem[{Liu et~al.(2019)Liu, Gao, Celikyilmaz, Carin
  et~al.}]{liu2019cyclical}
Xiaodong Liu, Jianfeng Gao, Asli Celikyilmaz, Lawrence Carin, et~al. 2019.
\newblock Cyclical annealing schedule: A simple approach to mitigating kl
  vanishing.
\newblock In \emph{Proceedings of the 17th Annual Conference of the North
  American Chapter of the Association for Computational Linguistics: Human
  Language Technologies (NAACL-HLT 2019)}.

\bibitem[{Miao and Blunsom(2016)}]{miao2016language}
Yishu Miao and Phil Blunsom. 2016.
\newblock Language as a latent variable: Discrete generative models for
  sentence compression.
\newblock In \emph{Proceedings of the 2016 Conference on Empirical Methods in
  Natural Language Processing}, pages 319--328.

\bibitem[{Miao et~al.(2016)Miao, Yu, and Blunsom}]{miao2016neural}
Yishu Miao, Lei Yu, and Phil Blunsom. 2016.
\newblock Neural variational inference for text processing.
\newblock In \emph{International conference on machine learning}, pages
  1727--1736.

\bibitem[{Park et~al.(2018)Park, Cho, and Kim}]{park2018hierarchical}
Yookoon Park, Jaemin Cho, and Gunhee Kim. 2018.
\newblock A hierarchical latent structure for variational conversation
  modeling.
\newblock In \emph{Proceedings of the 2018 Conference of the North American
  Chapter of the Association for Computational Linguistics: Human Language
  Technologies, Volume 1 (Long Papers)}, pages 1792--1801.

\bibitem[{Razavi et~al.(2019)Razavi, Oord, Poole, and
  Vinyals}]{razavi2019preventing}
Ali Razavi, A{\"a}ron van~den Oord, Ben Poole, and Oriol Vinyals. 2019.
\newblock Preventing posterior collapse with delta-vaes.
\newblock In \emph{International Conference on Learning Representations}.

\bibitem[{Semeniuta et~al.(2017)Semeniuta, Severyn, and
  Barth}]{semeniuta2017hybrid}
Stanislau Semeniuta, Aliaksei Severyn, and Erhardt Barth. 2017.
\newblock A hybrid convolutional variational autoencoder for text generation.
\newblock In \emph{Proceedings of the 2017 Conference on Empirical Methods in
  Natural Language Processing}, pages 627--637.

\bibitem[{Wang and Wang(2019)}]{wang2019riemannian}
Prince~Zizhuang Wang and William~Yang Wang. 2019.
\newblock Riemannian normalizing flow on variational wasserstein autoencoder
  for text modeling.
\newblock In \emph{Proceedings of the 2019 Conference of the North American
  Chapter of the Association for Computational Linguistics: Human Language
  Technologies, Volume 1 (Long and Short Papers)}, pages 284--294.

\bibitem[{Wen et~al.(2017)Wen, Miao, Blunsom, and Young}]{wen2017latent}
Tsung-Hsien Wen, Yishu Miao, Phil Blunsom, and Steve Young. 2017.
\newblock Latent intention dialogue models.
\newblock In \emph{Proceedings of the 34th International Conference on Machine
  Learning-Volume 70}, pages 3732--3741. JMLR. org.

\bibitem[{Xu and Durrett(2018)}]{xu2018spherical}
Jiacheng Xu and Greg Durrett. 2018.
\newblock Spherical latent spaces for stable variational autoencoders.
\newblock In \emph{Proceedings of the 2018 Conference on Empirical Methods in
  Natural Language Processing}, pages 4503--4513.

\bibitem[{Xu et~al.(2017)Xu, Sun, Deng, and Tan}]{xu2017variational}
Weidi Xu, Haoze Sun, Chao Deng, and Ying Tan. 2017.
\newblock Variational autoencoder for semi-supervised text classification.
\newblock In \emph{Thirty-First AAAI Conference on Artificial Intelligence}.

\bibitem[{Yang et~al.(2017)Yang, Hu, Salakhutdinov, and
  Berg-Kirkpatrick}]{yang2017improved}
Zichao Yang, Zhiting Hu, Ruslan Salakhutdinov, and Taylor Berg-Kirkpatrick.
  2017.
\newblock Improved variational autoencoders for text modeling using dilated
  convolutions.
\newblock In \emph{Proceedings of the 34th International Conference on Machine
  Learning-Volume 70}, pages 3881--3890. JMLR. org.

\bibitem[{Yang et~al.(2016)Yang, Yang, Dyer, He, Smola, and
  Hovy}]{yang2016hierarchical}
Zichao Yang, Diyi Yang, Chris Dyer, Xiaodong He, Alex Smola, and Eduard Hovy.
  2016.
\newblock Hierarchical attention networks for document classification.
\newblock In \emph{Proceedings of the 2016 Conference of the North American
  Chapter of the Association for Computational Linguistics: Human Language
  Technologies}, pages 1480--1489.

\bibitem[{Zhao et~al.(2017)Zhao, Zhao, and Eskenazi}]{zhao2017learning}
Tiancheng Zhao, Ran Zhao, and Maxine Eskenazi. 2017.
\newblock Learning discourse-level diversity for neural dialog models using
  conditional variational autoencoders.
\newblock In \emph{Proceedings of the 55th Annual Meeting of the Association
  for Computational Linguistics (Volume 1: Long Papers)}, pages 654--664.

\end{thebibliography}
\bibliographystyle{acl_natbib}


\end{document}


\appendix

\section{The Caveat about Implicit VAEs}
\label{app:ivae_caveat}

In this section, we show why the negative ELBO used as objective function and evaluation criteria from
\citet{fang2019implicit} is a lower bound on the negative ELBO of typical VAEs. The implication is that the estimated NLL
from \citet{fang2019implicit} cannot be directly compared with previous methods.
We provide an overview of the work by \citet{fang2019implicit},
followed with a detailed analysis of the caveat regarding their method
and why their experimental results for NLL are not to be directly compared to previous works.

The model proposed by \citet{fang2019implicit} is a variation of the sequence VAEs
that utilizes implicit distributions as their choice for variational posteriors
instead of the commonly used Gaussian distributions. The key idea is rooted on
the dual form of KL divergence based on the Fenchel duality theorem
\cite{dai2018coupled}:
\begin{align*}
    &D_{KL}(q_{\phi}(z|x)|p(z)) \\
    &= \max_{v \in \mathcal{F}_{+}} \mathbb{E}_{z \sim q_{\phi}}[v(x, z)]
                                    - \mathbb{E}_{z \sim p}[exp(v(x, z))] + 1
\end{align*}
where $v(x, z)$ is an auxiliary dual function in function space $\mathcal{F}_{+}$ which
contains \textit{all} positive functions.

The dual form would enable us to compute the KL term using samples from the approximate
posterior and the prior rather than analytically in the Gaussian case.
\citet{fang2019implicit} reported experimental results on text modelling which improved
upon the previous state-of-the-art by very large margins on standard benchmark datasets.
Please refer to Section 5.1 of their paper for more details.

The caveat is that the equality between the true $D_{KL}(q_{\phi}(z|x)|p(z))$ and its
dual form only holds if the dual function was the optimal one in the defined function
space. For their implementation, \citet{fang2019implicit} use a fixed-capacity neural
network to parameterize the dual function $v(x, z)$, which only covers a small subset
of the function space. This automatically renders the corresponding $\overline{D}_{KL}(q_{\phi}(z|x)|p(z))$
computed using the dual form a \textit{lower bound} on the true value of the KL term.
Additionally, their parameterized $v(x, z)$ is jointly trained with the rest of the model;
in practice there is no guarantee that the optimization would be able to find the
optimal function in even this small subset of the function space.

Therefore we have the following inequality:
\begin{align*}
    \overline{D}_{KL}(q_{\phi}(z|x)|p(z)) \leq D_{KL}(q_{\phi}(z|x)|p(z))
\end{align*}
Adding the reconstruction error to both sides:
\begin{align*}
    -\overline{\mathcal{L}}(x)
    &= -\mathbb{E}_{q_{\phi}(z|x)}[\log p_{\theta}(x|z)] + \overline{D}_{KL}(q_{\phi}|p) \\
    &\leq -\mathbb{E}_{q_{\phi}(z|x)}[\log p_{\theta}(x|z)] + D_{KL}(q_{\phi}|p) \\
    &= -\mathcal{L}(x)
\end{align*}
In other words, the negative ELBO $-\overline{\mathcal{L}}(x)$ that we obtain using
the dual form is in fact also a lower bound on the true negative ELBO
$-\mathcal{L}(x)$. In the most extreme case, $-\overline{\mathcal{L}}(x)$ could even
be sitting below the true negative log-likelihood of the data, which would be
problematic as it is meaningless to minimize a lower bound on the NLLs.

The same reasoning also applies to their evaluation. Since the results claimed in
\citet{fang2019implicit} were computed using the learned dual function $v$, their
reported negative ELBOs and in turn their estimated $\overline{\textbf{NLL}}$s are
a lower bound on the true \textbf{NLL}s of the data, thus
$\overline{\textbf{NLL}} \leq \textbf{NLL}$. Given the tools that we currently
have in learn theory, it is not trivial, if not impossible, to quantify the exact
gap between this lower bound and the true negative log-likelihoods. Therefore it
is unfair to compare their reported lower bound on NLLs with the exact results of
other models that follow the Gaussian assumption with analytical solutions to the
KL terms.

We choose to present the reported results of implicit VAEs for the sake of completeness
in terms of comparison. However we would like to point out this caveat of their method
and their evaluation so that progress that is made in this direction can be assessed
appropriately.

\bibliography{references}
\bibliographystyle{acl_natbib}